\documentclass[lettersize,journal]{IEEEtran}
\usepackage{amsmath,amsfonts}
\usepackage{algorithmic}
\usepackage{algorithm}
\usepackage{array}
\usepackage[caption=false,font=normalsize,labelfont=sf,textfont=sf]{subfig}
\usepackage{textcomp}
\usepackage{stfloats}
\usepackage{url}
\usepackage{verbatim}
\usepackage{graphicx}
\usepackage{cite}
\usepackage{booktabs}
\usepackage{tabularx}
\usepackage[colorlinks=true, urlcolor=blue, linkcolor=red]{hyperref}
\usepackage{hyperref}
\makeatletter
\def\UrlAlphabet{%
      \do\a\do\b\do\c\do\d\do\e\do\f\do\g\do\h\do\i\do\j%
      \do\k\do\l\do\m\do\n\do\o\do\p\do\q\do\r\do\s\do\t%
      \do\u\do\v\do\w\do\x\do\y\do\z\do\A\do\B\do\C\do\D%
      \do\E\do\F\do\G\do\H\do\I\do\J\do\K\do\L\do\M\do\N%
      \do\O\do\P\do\Q\do\R\do\S\do\T\do\U\do\V\do\W\do\X%
      \do\Y\do\Z}
\def\UrlDigits{\do\1\do\2\do\3\do\4\do\5\do\6\do\7\do\8\do\9\do\0}
\g@addto@macro{\UrlBreaks}{\UrlOrds}
\g@addto@macro{\UrlBreaks}{\UrlAlphabet}
\g@addto@macro{\UrlBreaks}{\UrlDigits}
\makeatother
\begin{document}

\title{Large Language Models for Robotics: Opportunities, Challenges, and Perspectives}


\author{Jiaqi Wang$^*$, Zihao Wu$^*$\thanks{$^*$Co-first authors.}, Yiwei Li, Hanqi Jiang, Peng Shu, Enze Shi, Huawen Hu, Chong Ma, Yiheng Liu, Xuhui Wang, Yincheng Yao, Xuan Liu, Huaqin Zhao, Zhengliang Liu, Haixing Dai, Lin Zhao,\\Bao Ge, Xiang Li, Tianming Liu$\dagger$, and Shu Zhang$\dagger$
\thanks{$\dagger$Co-corresponding authors: Tianming Liu, Shu Zhang}
\thanks{Jiaqi Wang, Enze Shi, Huawen Hu, Xuhui Wang, Yincheng Yao, and Shu Zhang are with the School of Computer Science, Northwestern Polytechnical University, Xi'an 710072, China. Chong Ma is with the School of Automation, Northwestern Polytechnical University, Xi'an 710072, China. (e-mail: \{jiaqi.wang, ezshi, huawenhu, xuhuiwang, yaoyincheng \}@mail.nwpu.edu.cn; shu.zhang@nwpu.edu.cn; mc-npu@mail.nwpu.edu.cn). }

\thanks{ Zihao Wu, Yiwei Li, Hanqi Jiang, Peng Shu, Huaqin Zhao, Zhengliang Liu, Haixing Dai, Lin Zhao, and Tianming Liu are with the School of Computing, The University of Georgia, Athens 30602, USA. Hanqi Jiang is with the College of Engineering, University of Georgia, Athens 30602, USA.(e-mail: \{zihao.wu1, yl80817, hj67104, peng.shu, hz33227, zl18864, haixing.dai, lin.zhao, tliu\}@uga.edu.)}

\thanks{Yiheng Liu and Bao Ge are with the School of Physics and Information Technology, Shaanxi Normal University, Xi’an
710119 China. Xuan Liu is with the School of Computer Science, Shaanxi Normal University, Xi’an 710119, China. (e-mail: \{liuyiheng, bob\_ge, xuanliu\}@snnu.edu.cn)}
\thanks{Xiang Li is with the Department of Radiology, Massachusetts General Hospital and Harvard Medical School, Boston 02115, USA. (e-mail: XLI60@mgh.harvard.edu).} 

}
\markboth{Journal of \LaTeX\ Class Files,~Vol.~14, No.~8, August~2021}%
{Shell \MakeLowercase{\textit{et al.}}: A Sample Article Using IEEEtran.cls for IEEE Journals}

\maketitle

\begin{abstract}
Large language models (LLMs) have undergone significant expansion and have been increasingly integrated across various domains. Notably, in the realm of robot task planning, LLMs harness their advanced reasoning and language comprehension capabilities to formulate precise and efficient action plans based on natural language instructions. However, for embodied tasks, where robots interact with complex environments, text-only LLMs often face challenges due to a lack of compatibility with robotic visual perception. This study provides a comprehensive overview of the emerging integration of LLMs and multimodal LLMs into various robotic tasks. Additionally, we propose a framework that utilizes multimodal GPT-4V to enhance embodied task planning through the combination of natural language instructions and robot visual perceptions. Our results, based on diverse datasets, indicate that GPT-4V effectively enhances robot performance in embodied tasks. This extensive survey and evaluation of LLMs and multimodal LLMs across a variety of robotic tasks enriches the understanding of LLM-centric embodied intelligence and provides forward-looking insights toward bridging the gap in Human-Robot-Environment interaction.
\end{abstract}

\begin{IEEEkeywords}
Large language model, robotic, GPT-4V, artifical general intelligence.
\end{IEEEkeywords}

\section{Introduction}
\IEEEPARstart{A}{s} pre-trained models have expanded in both model size and data volume, some large pre-trained models have demonstrated remarkable capabilities across a spectrum of complex tasks~\cite{zhao2023survey,wang2023prompt}. Large language models (LLMs) have garnered widespread attention in various domains due to their exceptional contextual emergence abilities~\cite{liu2023summary,liu2024understanding,zhou2023comprehensive,zhao2023brain,liu2022survey,wang2023prompt,rothman2022transformers,rahaman2023chatgpt,kaddour2023challenges}. This emergent capability empowers artificial intelligence algorithms in unprecedented ways, reshaping the manner in which people utilize artificial intelligence algorithms and prompting a reevaluation of the possibilities of Artificial General Intelligence (AGI).

With the rapid development of LLMs, the utilization of instruction tuning and alignment tuning has become the primary approach to adapt them for specific objectives. In the field of natural language processing (NLP), LLMs can, to some extent, function as a versatile solution for language-related tasks~\cite{liu2023summary,zhou2023comprehensive,liu2023radiology,ma2023impressiongpt,liu2023evaluating}. Research paradigms have also shifted towards employing LLMs to address subdomain-specific issues. In the realm of computer vision (CV), researchers are also working on developing large models, akin to GPT-4 and Gemini~\cite{openaiIntroducingChatGPT,team2023gemini}, that incorporate both visual and language information, thus supporting multimodal inputs~\cite{wang2023review}. This strategy of enhancing LLMs not only boosts their performance in downstream tasks but also holds significant guidance for the development of robotics by ensuring alignment with human values and preferences.

The ability of LLMs to process and internalize vast amounts of textual data offers unprecedented potential for enhancing a machine's understanding and natural language analysis capabilities~\cite{khurana2023natural,li2023data}. This extends to comprehending documents like manuals and technical guides and applying this knowledge to engage in coherent, accurate, and human-aligned dialogues~\cite{huang2023free,liu2023llmp,wang2023aligning}. Through conversation, natural language instructions are translated from text prompts into machine-understandable code that triggers corresponding actions, thereby rendering robots more adaptive and flexible in accommodating a wide array of user commands~\cite{yoneda2023statler,liang2023code,vemprala2023chatgpt}. Integrating real-world sensor modalities into language models facilitates establishing connections between words and perceptions, enabling their application across various specific tasks. Nevertheless, text only LLMs lack experiential exposure to the physical world and the empirical outcomes of observation, making it challenging to employ them in decision-making within specific environments. Therefore, incorporating multimodality into LLMs is crucial for the effective execution of robotic tasks. Additionally, the field of robotics presents subtler variations in tasks. Unlike NLP and CV, which can leverage extensive datasets from the internet, acquiring large and diverse datasets for robot interactions is challenging~\cite{padalkar2023open}. These datasets often either focus on a single environment and object or emphasize specific task domains, resulting in substantial differences between them.~\cite{mai2023llm} This intricacy presents more significant challenges when integrating LLMs with robotics.

How to overcome the challenges posed by robotic technology and harness the accomplishments of LLMs in other domains for the benefit of the robotics field is the central inquiry addressed in this review. In this article, the work’s contributions can be summarized in four main points.
\begin{itemize}
\item We meticulously survey and synthesize existing LLM for robotic literature, exploring the latest advancements in three distinct task categories: planning, manipulation, reasoning.
\item We summarize the primary technical approaches that LLMs offer to the realm of robotics, examine the potential for training generalized robot strategies, and provide a foundational survey for researchers in this domain. 
\item We assess the effectiveness of multimodal GPT-4V in robot task planning across various environments and scenarios. 
\item We summarize the key findings of our investigation, deliberate upon the outstanding challenges to be tackled in future endeavors, and present a forward-looking perspective.
\end{itemize}







\begin{figure*}[t]
\centering
\includegraphics[width=0.8\textwidth]{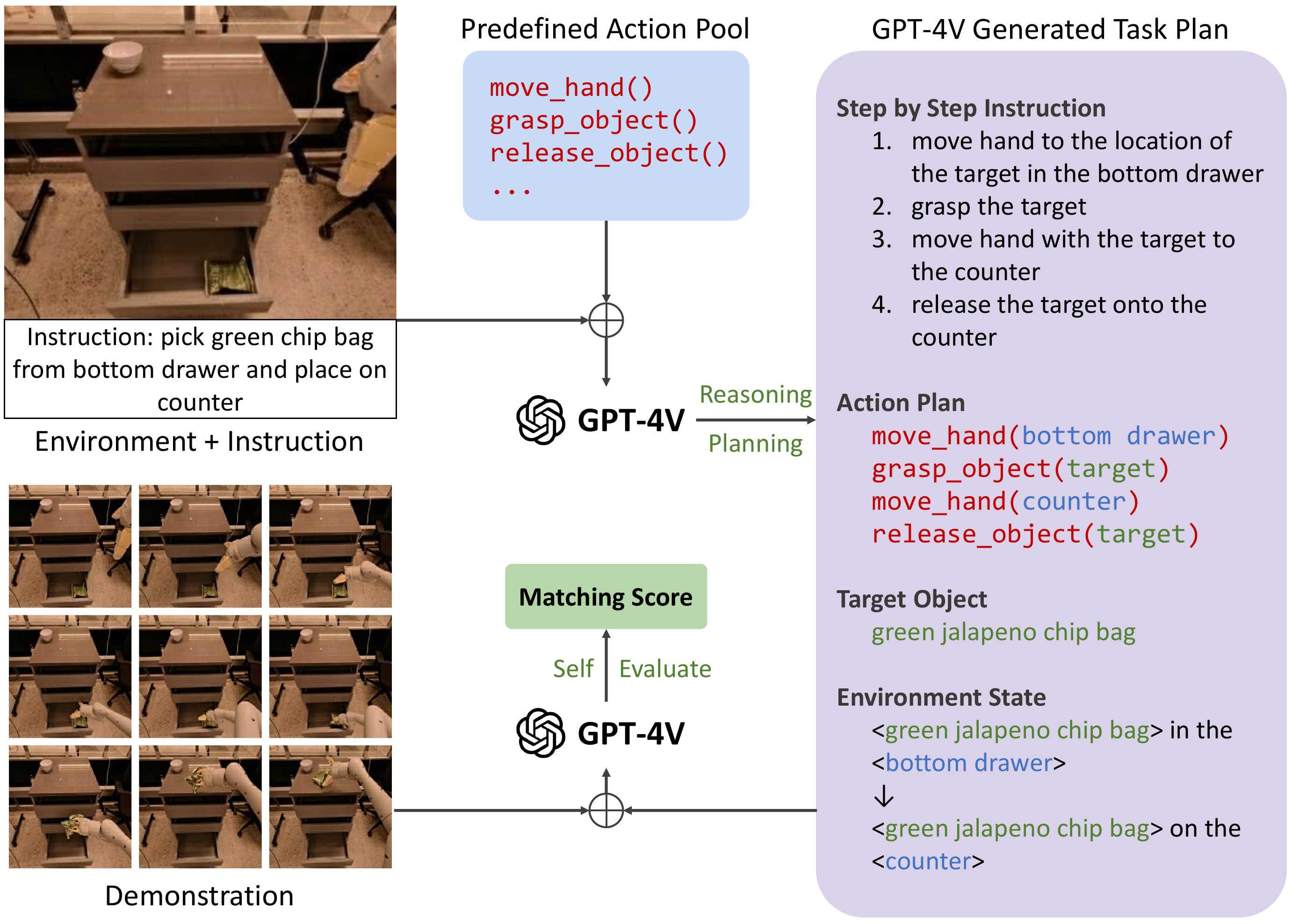}
\caption{Framework of the proposed GPT-4V empowered embodied task planning. We use the initial frame of video data along with their corresponding textual instructions as input. Our framework, leveraging GPT-4V, breaks down the instructions into a sequence of task plans and selects corresponding representations from a predefined action pool. Simultaneously, we can analyze the target object related to the instruction and the environmental changes before and after the instruction in the images. Finally, we employ GPT-4V to compare and score the task plan we generated against the ground truth plan.} 
\label{main}
\end{figure*}

\section{Related Work}

\subsection{LLM for Robotics}
The field of robotics research based on LLMs has made significant strides. These models demonstrate exceptional natural language understanding and commonsense reasoning capabilities, significantly enhancing a robot's ability to comprehend contexts and execute commands. Current research focuses on leveraging LLMs to parse complex contexts and instructions, including addressing ambiguity, resolving ambiguities, and understanding implicit information. A key advancement in this domain includes the development of vision-language models,~\cite{zhang2021vinvl,zhou2022learning,zhu2023minigpt} which have markedly improved the performance of tasks like visual question answering~\cite{lin2023medical,ravi2023vlc,shao2023prompting} and image captioning.~\cite{brooks2023instructpix2pix,gorokhovatskyi2023search} These advancements have greatly boosted a robot's ability to reason in the physical world, particularly in areas such as complex command navigation.~\cite{shah2023lm,huang2023visual} Through visual language processing systems, robots are capable of understanding image content and integrating it with relevant linguistic information, such as image descriptions and command execution. This multimodal information processing is similarly applied in audio-visual integration. Another major progress with LLMs is in human-robot interaction, facilitated by interactive learning processes that better align with human needs and preferences. For example, by integrating reinforcement learning with human feedback, robots can continuously improve their task execution, addressing semantic ambiguities encountered in large model applications, by combining human guidance with large language models, robots can refine instructions more precisely, thereby better achieving autonomous learning and environmental adaptation for more accurate and targeted control. Robots can also learn and adapt to user behavior, preferences, and needs through interaction, providing a more personalized and customized interaction experience. These advancements not only enhance the practicality of robotic technology but also open up new possibilities for future human-machine interactions.

\subsection{Multimodal Task Planning with LLMs}

Multimodal Tasks Planning within the domain of LLMs constitutes a sophisticated intersection of artificial intelligence disciplines, engaging an amalgamation of disparate data modalities — such as textual, visual, and auditory inputs — to foster a more holistic and nuanced AI-driven analysis~\cite{brohan2023can,brohan2023rt,cui2023no,huang2023voxposer,huang2022inner}.

This interdisciplinary approach transcends the traditional boundaries of LLMs, which predominantly focused on textual comprehension and generation, ushering in an era where these models are adept at interpreting, correlating, and interacting with multiple data streams in unison. In this context, the LLM's role evolves from mere language processing to a more integrative function, synthesizing and responding to complex data interplays. In the realm of Multimodal Tasks Planning with LLMs, recent advancements exemplified by projects like Inner Monologue and SayCan demonstrate the burgeoning complexity and sophistication in this field. Inner Monologue's~\cite{huang2022inner} methodology represents a significant leap in this domain, as it integrates multi-modal feedback sources from the environment. This integration enables the generation of more reliable and contextually aware task planning, harmonizing different sensory inputs to create a more cohesive understanding of the AI's surroundings. Similarly, the SayCan's~\cite{brohan2023can} framework introduces a novel dimension to LLM applications. This system employs LLMs as a proxy for the model’s “hands and eyes,” generating the optimal long-horizon instructions and effectively scoring the affordance probability of the instruction on the current scene. This methodology not only enhances the AI's ability to understand and interact with its immediate environment but also leverages the nuanced understanding of LLMs to plan and execute complex sequences of actions over extended periods.

The integration of these advanced techniques in Inner Monologue and SayCan within multimodal task planning with LLMs represents a significant stride towards creating AI systems that are not only more cognizant of multiple data streams but are also capable of synthesizing these streams into actionable intelligence. This progression points towards a future where AI can navigate and interact with the real world in a manner that is far more dynamic, context-aware, and autonomous~\cite{brohan2023can,huang2022inner,zhao2023large,zeng2022socratic}, pushing the boundaries of what is achievable in AI-driven innovation and interdisciplinary synthesis.


\section{Scope of Robotic Tasks}

\subsection{Planning}

\subsubsection{Natural Language Understanding }

In robot planning, Large Language Models excel due to their advanced natural language comprehension. They translate natural language instructions into executable action sequences for robots, a crucial aspect of robot planning~\cite{driess2023palm,brohan2023can}. This study reveals that LLMs can generate accurate action sequences based on linguistic instructions alone, even without visual input~\cite{abdulsaheb2023classical}. However, their performance is enhanced significantly with a modest amount of visual information, enabling them to create precise visual-semantic plans. These plans transform high-level natural language instructions into actionable guidance for virtual agents to undertake complex tasks. This ability underscores the potential of LLMs to integrate multimodal information, thereby improving their comprehension. It also demonstrates their capacity to interpret and incorporate information from various modalities, leading to a more comprehensive task understanding~\cite{jansen2020visually}. Moreover, research in generating action sequences from a large language model for natural language understanding further confirms the efficacy of LLMs in robot planning. LLMs also show great promise in interpreting natural language commands in sync with the physical environment. Employing the Grounded Decoding approach, they can produce behavior sequences that align with the probabilities of the physical model, showcasing this method's effectiveness in robot planning tasks~\cite{huang2023grounded}.


The research in complex sequential task planning has highlighted significant advancements in the capabilities of LLMs. Text2Motion's studies demonstrate that LLMs are adept not only at processing linguistic information but also at addressing dependencies in skill sequences~\cite{lin2023text2motion}. This is achieved through geometrically feasible planning, marking a crucial advancement in the interpretation of abstract instructions and the comprehension of intricate task structures. In addition, LLM-Planner research enhances the natural language understanding abilities of LLMs in robotic planning by integrating them with conventional planners~\cite{song2023llm}. This synergy illustrates how the NLP proficiencies of LLMs can be harnessed to boost the efficiency and precision of planning tasks. Moreover, LLM+P harnesses the capabilities of classical planners, employing the Planning Domain Definition Language (PDDL) and problem cues to create task-specific problem files for LLMs~\cite{liu2023llmp}. This integration significantly amplifies LLMs' efficacy in addressing long-term planning tasks. Also, SayPlan addresses the issue of planning horizon by integrating a classical path planner. By doing so, SayPlan is capable of grounding large-scale, long-horizon task plans derived from abstract and natural language instructions, enabling a mobile manipulator robot to execute them successfully~\cite{rana2023sayplan}. Furthermore, LLMs have shown promise in acting as heuristic strategies within search algorithms, while also serving as reservoirs of common-sense knowledge. This dual role of LLMs not only enhances the reasoning capabilities within these algorithms but also aids in forecasting potential outcomes. Such an approach harnesses the full potential of LLMs, leveraging their advanced reasoning capabilities for the effective planning of complex tasks~\cite{zhao2023large}. This dual application underscores the extensive and versatile potential of large language models in task planning and problem-solving.

The research conducted on LLMs has showcased their remarkable ability to parse and comprehend natural language understanding. This capability extends beyond mere text matching to a profound semantic understanding, encompassing the tasks' purpose and context. A critical aspect of LLMs is translating the instructions they comprehend into executable action sequences for robots, an essential feature in robot task planning. LLMs significantly enhance the quality and adaptability of instruction generation, enabling the creation of complex action sequences that are both context-aware and environment-specific. These models demonstrate versatility in managing various task-planning complexities and types, from straightforward physical interactions to intricate, long-term sequence planning. The studies highlight LLMs' potential as both independent decision-makers and collaborators with other modalities and planning algorithms. This collaboration is pivotal in interpreting natural language and advancing robotic planning. As research progresses, LLMs are expected to play an increasingly vital role in the fields of robotics and automated systems.



\subsubsection{Complex Task Reasoning and Decision-making}

In the realm of complex task reasoning and decision-making, robots empowered by LLMs have shown remarkable proficiency. These LLM-based robotic planning tasks have significantly transcended the realms of mere text generation and language comprehension. Recent research highlights the immense capabilities of Language Models in managing intricate tasks, engaging in logical reasoning, making informed decisions, and partaking in interactive learning.~\cite{chang2023survey,liu2023summary} These breakthroughs have not only expanded our comprehension of LLM-based robotic planning's potential but also opened the door to innovative practical applications.

In exploring the application of pre-trained language models (PLMs) in interactive decision-making, research has demonstrated how targets and observations are transformed into embedding sequences, initializing the network with PLMs. This strategy's generalization ability is particularly effective in multivariate environments and supervised modalities~\cite{li2022pre}. A notable advancement in the multimodal domain is the development of the LM-Nav system~\cite{shah2023lm}. This system, grounded in PLMs, integrates language, vision, and action models to guide robotic navigation via high-level natural language commands. Significantly, it reduces dependency on costly trajectory annotation supervision by merging pre-trained visual navigation, image-verbal correlation, and language understanding models. Focusing on LLMs in specific environments, researchers~\cite{huang2022inner} have examined their capacity for reasoning with natural language feedback and complex task planning. This capability is crucial for following high-level task instructions and enhancing the model's applicability in real-world scenarios. Addressing the issue of consistency fault-tolerance in natural language understanding and decision-making, the innovative ReAct model~\cite{yao2022react} overcomes prior limitations of linguistic reasoning in interactive settings. It tackles challenges like hallucination generation and misinformation propagation. By leveraging LLMs' potential to maintain working memory and abstractly conceptualize high-level goals, the ReAct model achieves significant performance improvements across various tasks. In parallel, to address confidently hallucinated predictions in large language models (LLMs) applied to robotics, KnowNo~\cite{ren2023robots} provides statistical guarantees for task completion while minimizing the need for human assistance in complex multi-step planning scenarios. Notably, KnowNo seamlessly integrates with LLMs without requiring model-finetuning, offering a lightweight and promising method to model uncertainty. This approach aligns with the constantly evolving capabilities of foundation models, providing a scalable solution. Further, a strategy involving preconditioned error cues has been proposed, enabling LLMs to extract executable plans. This approach offers a fresh perspective on the independence and adaptability of agents in task execution. In terms of multi-agent collaboration, the integration of language models with action agents is increasingly being explored. By pairing LLMs with agents executing tasks in specific environments, a system comprising planners, executors, and reporters is established. This arrangement markedly enhances the efficiency of reasoning and execution in complex tasks.

The burgeoning field of large pre-trained LMs is witnessing a notable trend: these models are increasingly adept at understanding and performing complex tasks, closely aligning with real-world scenarios. This advancement not only underscores the adaptability and versatility of pre-trained models but also heralds the advent of next-generation AI. As these technologies evolve, we anticipate a surge in innovative applications, poised to revolutionize various industries. A key aspect of these tasks is the utilization of LLMs' robust language comprehension and generation capabilities for intricate reasoning and decision-making processes. Each study in this domain explores the potential of LLMs in complex cognitive functions. Many models employ self-supervised learning, with some incorporating fine-tuning to better align with specific tasks. This approach enables LLMs to excel in downstream task-assisted reasoning, leading to more precise and tailored decisions. Despite the widespread use of LLMs in complex reasoning and decision-making, the specific techniques and approaches vary, particularly in terms of task handling, learning strategies, and feedback mechanisms. These models find applications in diverse real-world contexts, including home automation, robot navigation, and task planning, demonstrating their broad and evolving utility.

\subsubsection{Human-robot interaction}

In the realm of human-robot interaction, the advanced reasoning capabilities of AGI language models empower robots with a significant degree of generalization ability.~\cite{rane2023transformers} This enables them to adapt to new task planning in previously unseen environments and tasks. Furthermore, the natural language understanding interface of LLMs facilitates communication with humans, opening new possibilities for human-robot interactions.~\cite{qiao2023brain} Extensive research has underscored the progress made by LLMs in aiding intelligent task planning, which in turn enhances multi-intelligence collaborative communication. Studies have found that using natural language to boost the efficiency of multi-intelligence cooperation is an effective method to enhance communication efficiency. A notable example of this is OpenAI's ChatGPT, whose capabilities in robotics applications were evaluated through rigorous experiments. The findings revealed that ChatGPT excels in complex tasks such as logical, geometric, and mathematical reasoning, along with airborne navigation, manipulation, and controlling embodied agents~\cite{vemprala2023chatgpt}. It achieves this through techniques like free-form dialogue, parsing XML tags, and synthesizing code. Furthermore, ChatGPT allows user interaction via natural language commands, providing vital guidance and insights for the development of innovative robotic systems that interact with humans in a natural and intuitive way. In a similar vein, there is a proposed framework that leverages large-scale language models for collaborative embodied intelligence~\cite{zhang2023building}. This framework enables the use of language models for efficient planning and communication, facilitating collaboration between various intelligences and humans to tackle complex tasks. Experimental results demonstrate that this approach significantly outperforms traditional methods in the field.



\subsection{Manipulation}

\subsubsection{Natural Language Understanding}

In the field of robot control, the natural language understanding capabilities of LLM can help robots make common-sense analyses. For example, LLM-GROP demonstrates how semantic information can be extracted from LLM and used as a way to make common-sense, semantically valid decisions about object placement as part of a task and motion planner that performs multistep tasks in complex environments in response to natural language commands~\cite{ding2023task}. The research proposes a framework for placing language at the core of an intelligent body~\cite{di2023towards}. By utilizing the prior knowledge contained in these models, better robotic agents can be designed that are able to solve challenging tasks directly in the real world. Through a series of experiments, it is demonstrated how the framework can be used to solve a variety of problems with greater efficiency and versatility by utilizing the knowledge and functionality of the underlying models. At the same time, the study introduces Linguistically Conditional Collision Function (LACO), a novel method to learn collision functions using only single-view image, language prompt, and robot configuration. LACO predicts collisions between robots and the environment, enabling flexible conditional path planning~\cite{xie2023language}.



Outside of natural language understanding capabilities, the powerful reasoning capabilities of LLM also have a prominent role. For example, in the VIMA work~\cite{jiang2022vima}, a novel multimodal cueing formulation is introduced to convert different robot manipulation tasks into a unified sequence modeling problem and instantiated in a diverse benchmark with multimodal tasks and system generalization evaluation protocols. Experiments show that VIMA is capable of solving tasks such as visual goal realization, one-off video imitation, and novel conceptual foundations using a single model, with robust model scalability and zero-sample generalization. Similarly, TIP proposes Text-Image Cueing~\cite{lu2023multimodal}, a bimodal cueing framework that connects LLMs to multimodal generative models for rational multimodal program plan generation.

In addition to prompt methods, fine-tuning downstream tasks based on pre-trained LMs is also a common approach in the field of robot control. For example, the work demonstrated that pre-trained visual language representations can effectively improve the sample efficiency of existing exploratory methods~\cite{tam2022semantic}. R3M investigates how pre-trained visual representations on different human video data can enable data-efficient learning of downstream robot manipulation tasks~\cite{nair2022r3m}. The LIV is trained on a large generalized human video dataset, and fine-tuned on a small robot data set, is fine-tuned to outperform state-of-the-art methods in three different evaluation settings, and successfully performs real-world robotics tasks~\cite{ma2023liv}.

This collection of studies collectively illustrates the significant role of LLMs and Natural Language Understanding techniques in advancing robotic intelligence, particularly in comprehending and executing complex, language-based tasks. A key emphasis of these studies is on the importance of model generalization and the ability to apply these models across various domains. Each study, while sharing this common theme, diverges in its specific focus and application methodology. For instance, LLM-GROP is dedicated to the extraction and application of semantic information. In contrast, VIMA and TIP concentrate on multimodal processing and learning without prior examples. Furthermore, methodologies that fine-tune pre-trained LMs are directed toward enhancing application efficiency and task-specific optimization. Collectively, these studies demonstrate that integrating sophisticated NLP techniques with machine learning strategies can substantially enhance the efficiency of robotic systems, particularly in their ability to understand and perform intricate tasks. This advancement is a crucial stride towards achieving greater intelligence and autonomy in robotic manipulation.



\subsubsection{Interactive Strategies}

In the realm of interactive strategies, the TEXT2REWARD framework introduces an innovative approach for generating interactive reward codes using LLMs~\cite{di2023towards}. This method automatically produces dense reward codes, enhancing reinforcement learning. Also, By utilizing Large Language Models to define reward parameters that can be optimized to accomplish a variety of robotic tasks, the gap between high-level language instructions or corrections and low-level robot actions can be effectively bridged. The rewards generated by the language models serve as an intermediate interface, enabling seamless communication and coordination between high-level instructions and low-level actions of the robot~\cite{yu2023language}. Furthermore, VoxPoser presents a versatile framework for robot manipulation~\cite{huang2023voxposer}, distinct in its ability to extract manipulability and constraints directly from LLMs. This approach significantly enhances the adaptability of robots to open-set instructions and diverse objects. By integrating LLMs with vision-language models and leveraging online interactions, VoxPoser efficiently learns to interact with complex task dynamics models. The application of LLMs extends to human-robot interaction as well. The LILAC system exemplifies this through a scalable~\cite{cui2023no}, language-driven interaction mechanism between humans and robots. It translates natural language discourse into actionable commands within a low-dimensional control space, enabling precise and user-friendly guidance of robots. Importantly, each user correction refines this control space, allowing for increasingly targeted and accurate commands. InstructRL offers another innovative framework designed to enhance human-AI collaboration~\cite{hu2023language}. It focuses on training reinforcement learning agents to interpret and act on natural language instructions provided by humans. This system employs LLMs to formulate initial policies based on these instructions, guiding reinforcement learning agents toward achieving an optimal balance in coordination. Lastly, for language-based human-machine interfaces, a novel, flexible interface LILAC has been developed. It permits users to alter robot trajectories using textual input and scene imagery~\cite{bucker2023latte}. This system synergizes pre-trained language and image models like BERT and CLIP, employing transformer encoders and decoders to manipulate robot trajectories in both 3D and velocity spaces. Proving effective in simulated environments, this approach has also demonstrated its practicality through real-world applications.

All of these techniques and approaches depend, to varying degrees, on advanced language modeling to enhance human-robot interaction and robot control. They collectively underscore the crucial role of LLMs in interpreting and executing human intentions. Each method aims to boost the adaptability and flexibility of robots, enabling them to handle diverse tasks and environments more effectively. Specifically, TEXT2REWARD centers on generating and optimizing reward codes. This enhances the efficacy of reinforcement learning strategies. Conversely, VoxPoser focuses on extracting operants and constraints from LLMs. Meanwhile, LILAC and InstructRL adopt distinct approaches to interpreting and executing natural language commands. LILAC prioritizes mapping discourse to a control space, whereas StructRL dedicates itself to training reinforcement learning agents to comprehend and follow natural language instructions. Additionally, the last discussed language-based Human-Machine Interaction research investigates how to directly extract user intentions from text and images, applying them across various robot platforms. This aspect sets it apart from other approaches that might not incorporate this feature. Collectively, these studies mark substantial advancements in integrating LLMs techniques into robotics. While their application areas and methodologies have distinct focal points, they collectively demonstrate the potential for innovation in artificial intelligence. Furthermore, they pave the way for future explorations in human-robot interaction.

\subsubsection{Modular Approaches}

Recent advancements in robot control emphasize modular approaches, allowing the creation of more complex and feature-rich robotic systems. Key aspects of this trend have been highlighted in recent research. PROGRAMPORT proposes a program-based modular framework focused on robot manipulation~\cite{wang2023programmatically}.  It interprets and executes linguistic concepts by translating natural language's semantic structure into programming elements. The framework comprises neural modules that excel in learning both general visual concepts and task-specific operational strategies. This structured approach distinctly enhances learning of visual foundations and operational strategies, improving generalization to unseen samples and synthetic environments.

Next, researchers have explored the use of LLMs to expedite strategy adaptation in robotic systems~\cite{ren2023leveraging}, particularly when encountering new tools. By generating geometrical shapes and descriptive tool models, and then converting these into vector representations, LLMs facilitate rapid adaptation. This integration of linguistic information and meta-learning has shown significant performance improvements in adapting to unfamiliar tools.
 
In addition, Combining NLMap~\cite{chen2023open}, a visual language model based on ViLD and CLIP, with the SayCan framework, has led to a more flexible scene representation. This combination is particularly effective for long-term planning, especially when processing natural language commands in open-world scenarios. NLMap enhances the capability of LLM-based planners to understand their environments.
 
The "Scaling Up and Distilling Down" framework combines the advantages of LLMs~\cite{ha2023scaling}, sampling-based planners, and policy learning. It automates the generation, labeling, and extraction of rich robot exploration experiences into a versatile visual-linguistic motion strategy. This multi-task strategy not only inherits long-term behavior and robust manipulation skills but also shows improved performance in scenarios outside the training distribution.

MetaMorph introduces a Transformer-based method for learning a generalized controller applicable to a vast modular robotic design space~\cite{gupta2022metamorph}. This approach enables the use of robot morphology as a Transformer model output. By pre-training on a diverse range of morphologies, strategies generated through this approach demonstrate broad generalizability to new morphologies and tasks. This showcases the potential for extensive pre-training and fine-tuning in robotics, akin to developments in vision and language fields.

In each of these studies, a modular approach has been adopted, enhancing the system's flexibility and adaptability to new tasks and environments. These works extensively utilize deep learning techniques, notably in synergy with LLMs, to augment the robotic system's understanding and decision-making abilities. Moreover, a significant focus of these studies is the application of NLP. This is evident either through the direct interpretation of linguistic commands or via linguistically enriched learning and adaptation processes. The primary objective is to enhance the robot's capability for quick generalization and adaptation in novel environments and tasks. While all the studies employ deep learning and LLMs, their specific implementations and applications are diverse. Some are centered on linguistic description and comprehension, while others explore the fusion of vision and language. The research goals are varied, addressing challenges from adapting to new tools, to long-term strategic planning, to polymorphic robot control. Despite differences in technical approaches, application areas, and targeted tasks, each study significantly contributes to advancing the intelligence and adaptive capabilities of robotic systems.

\subsection{Reasoning}

\subsubsection{Natural Language Understanding}

In the realm of robotic reasoning tasks, LLMs based on natural language understanding serve as an essential knowledge base, providing common sense insights crucial for various tasks. Extensive research has shown that LLMs effectively simulate human-like states and behaviors, especially relevant in the study of robots performing household cleaning functions. This approach deviates from traditional methods, which typically require costly data gathering and model training. Instead, LLMs leverage off-the-shelf methods for generalization in robotics, benefiting from their robust summarization abilities honed from extensive textual data analysis. Moreover, the common sense reasoning and code comprehension capabilities of LLMs foster connections between robots and the physical world. For instance, Progprompt introducing programming language features in LLMs has been shown to enhance task performance. This approach is not only intuitive but also sufficiently flexible to adapt to new scenarios, agents, and tasks, including actual robot deployments~\cite{singh2023progprompt}. Concurrently, GIRAF harnesses the power of large language models to more flexibly interpret gestures and language commands, enabling accurate inference of human intentions and contextualization of gesture meanings for more effective human-machine collaboration~\cite{lin2023gesture}.

One innovative development in this field is Cap (Code as Policies)~\cite{liang2023code}, which advocates for robot-centric language model generation programs. These programs can be adapted to specific layers of the robot's operational stack: interpreting natural language commands, processing perceptual data, and parameterizing low-dimensional inputs for the original language control. The underlying principle of this approach is that layered code generation facilitates the creation of more intricate code, thereby advancing the state-of-the-art in this area.

Both the home cleaning application and the robot-centric language model generation programs in Cap highlight the strengths of LLMs in providing common sense knowledge and interpreting natural language instructions. Traditional robotics often necessitates extensive data collection and specialized model training. In contrast, LLMs mitigate this need by utilizing their extensive training on textual data. The code comprehension and generation abilities of LLMs are particularly crucial, enabling robots to interact more effectively with the physical world and execute complex tasks. However, there is a distinction in application focus: the home cleaning function tends to emphasize everyday tasks and environmental adaptability, whereas Cap centers on programming and controlling the robot's more technical behaviors through Language Model Generation Programs (LMPs).

In summary, the integration of LLMs into robotic reasoning tasks underscores their remarkable capabilities in natural language understanding, common sense knowledge provision, and code comprehension and generation. These features not only alleviate the data collection and model training burdens typically associated with traditional robotics but also enhance robots' generalization and flexibility. With adequate training and adjustment, LLMs can be applied across various scenarios and tasks, demonstrating their vast potential and wide-ranging applicability in the future of robotics and artificial intelligence.

\subsubsection{Complex Task Reasoning and Decision-making}



In the realm of complex task reasoning and decision-making, various studies have leveraged the reasoning abilities of LLMs to augment the refinement of specific downstream tasks. For instance, SayCan utilizes the extensive knowledge embedded in LLMs for concretization tasks alongside reinforcement learning~\cite{brohan2023can}. This method involves using reinforcement learning to uncover insights about an individual's skill value function. It then employs textual labels of these skills as potential responses, while the LLM provides overarching semantic guidance for task completion.

Another notable development is the Instruct2Act framework~\cite{huang2023instruct2act}. It offers a user-friendly, general-purpose robotics system that employs LLMs to translate multimodal commands into a sequence of actions in the robotics field. This system uses policy code generated by LLMs, which make API calls to various visual base models, thus attaining a visual comprehension of the task set.

The usage of LLMs for self-planning and in PDDL (Planning Domain Definition Language) planning has also been explored~\cite{silver2022pddl}. It has been shown that LLM outputs can guide heuristic search planners effectively.

In the domain of failure explanation and correction tasks, the REFLECT framework leverages a hierarchical summary of the robot's past experiences generated from multisensory observations to query an LLM for failure reasoning~\cite{liu2023reflect}. The failure explanation obtained can then guide a language-based planner to correct the failure and successfully accomplish the task.

Furthermore, the adaptation of pre-trained multimodal models is a common strategy. By integrating the pre-training of vision-language models with robot data to train Visual-Linguistic-Action (VLA) models~\cite{brohan2023rt}, researchers have found that models trained on internet data with up to 55 billion parameters can generate efficient robot strategies. These models exhibit enhanced generalization performance and benefit from the extensive visual-linguistic pre-training capabilities available on the web.

Socratic Models represent another approach~\cite{zeng2022socratic}, where structured dialogues between multiple large pre-trained models facilitate joint predictions for new multimodal tasks. This method has achieved zero-shot performance across multiple tasks.

In these studies, the primary focus has been on harnessing LLMs for automating reasoning and decision-making processes. This is achieved by leveraging LLMs' capacity to provide or utilize high-level semantic knowledge, thereby enhancing task execution. Some approaches integrate LLMs with other modalities, like vision and action, to deepen task understanding and execution. Others demonstrate effective performance on previously unseen tasks, showcasing zero-shot or few-shot learning capabilities.

Each study adopts a unique approach to integrate LLMs. For example, SayCan incorporates reinforcement learning, whereas Instruct2Act is centered on the direct mapping of multimodal instructions. The techniques employed—ranging from reinforcement learning and heuristic search to multimodal pretraining—vary significantly across different application domains like robot manipulation, planning, and automated decision-making. These studies collectively illustrate LLMs' vast potential in managing complex task reasoning and decision-making. By amalgamating LLMs with other techniques, such as reinforcement learning and multimodal data processing, a deeper semantic understanding and more effective decision support can be achieved. This is particularly evident in robotics and automation, where such integrated approaches are paving the way for novel applications. However, the efficacy of these methods is highly contingent on the specific nature of the task, the data utilized, and the model training approach. Hence, the selection and application of each method must be meticulously tailored to the specific context.

\subsubsection{Interactive Strategies}
The recent advancements in LLMs have significantly contributed to the development of interactive strategies, showcasing impressive capabilities in language generation and human-like reasoning. Matcha~\cite{zhao2023chat}, utilizing LLMs, enhances interactive multimodal perception, illustrating the potential of LLMs in understanding various types of input data, such as visual and auditory. This approach proposes an augmented LLM multimodal interactive agent. This agent not only leverages commonsense knowledge inherent in LLMs for more plausible interactive multimodal perception but also demonstrates the practical application of LLMs in conducting such perception and interpreting behavior.

Generative agents, as introduced are interactive computational agents designed to simulate human behavior~\cite{park2023generative}. The architecture of these agents is engineered to store, synthesize, and apply relevant memories, thereby generating plausible behaviors using large language models. The integration of LLMs with these computational agents facilitates the creation of advanced architectures and interaction patterns. This combination enables more realistic simulations of human behavior, extending the potential applications of LLMs.

The emphasis in LLM-based interactive strategies is on the fusion of LLMs with other perceptual systems, such as image recognition and speech processing. This amalgamation aims to mimic or augment human abilities, enhancing cognitive and processing capabilities. Such advancements have profound implications in the realms of intelligent assistants, robotics, and augmented reality systems.

In the discussed work, a notable emphasis is placed on multimodal perception, focusing on improving the system's ability to understand and interact with its environment. Additionally, the simulation of human behavior seeks to replicate human thought and action processes in AI. The convergence of these two directions holds the promise of creating more powerful and versatile intelligent systems. These systems are envisioned to interact with humans at a more complex and humanized level, presenting significant technical challenges as well as raising crucial ethical and social adaptation questions.

\begin{table*}[h!]
\centering
\caption{Description of the dataset and the average matching scores self-evaluated by GPT-4V, comparing the task plans it generated with the ground truth demonstrations across nine tested datasets.}
\label{match_score}
\begin{tabular}{llc}
\toprule
Dataset & Description & Matching Score\\
\midrule
RT-1 Robot Action~\cite{brohan2022rt} & Robot picks, places and moves 17 objects from the google micro kitchens. & 9/10\\
QT-Opt~\cite{kalashnikov2018qt} & Kuka robot picking objects in a bin. & 8/10\\
Berkeley Bridge~\cite{walke2023bridgedata} & The robot interacts with household environments including kitchens, sinks, and tabletops. & 8.7/10\\
TOTO Benchmark~\cite{zhou2023train} & The TOTO Benchmark Dataset contains trajectories of two tasks: scooping and pouring. & 8.5/10\\
BC-Z~\cite{jang2021bc} & The robot attempts picking, wiping, and placing tasks on a diverse set of objects on a tabletop, & 7.7/10\\
Berkeley Autolab UR5~\cite{BerkeleyUR5Website}& The data consists of 4 robot manipulation tasks. & 8.3/10\\
NYU VINN~\cite{pari2021surprising}& The robot arm performs diverse manipulation tasks on a tabletop. & 9/10\\
Freiburg Franka Play~\cite{rosete2022tacorl} & The robot interacts with toy blocks, it pick and places them, stacks them, & 10/10\\
USC Jaco Play~\cite{dass2023jacoplay} & The robot performs pick-place tasks in a tabletop toy kitchen environment. & 9/10\\
All & - & 8.7/10\\
\bottomrule
\end{tabular}
\end{table*}

    



\begin{figure*}[h!]
\centering
\includegraphics[width=1\textwidth]{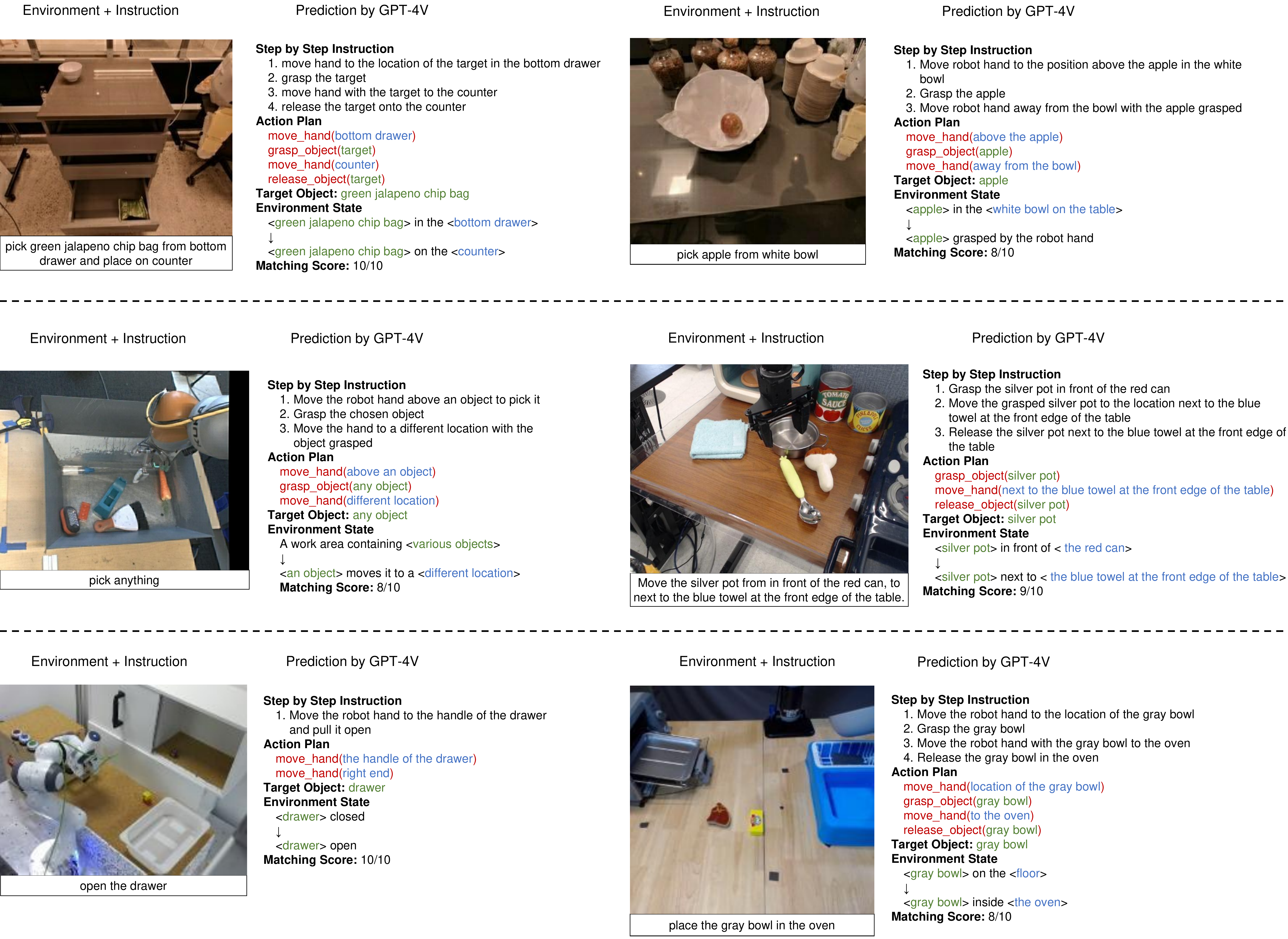}
\caption{Generated task plans for different datasets: RT-1 Robot Action (Top Panel), QT-Opt (Middle Left), Berkeley Bridge (Middle Right), Freiburg Franka Play (Bottom Left), and USC Jaco Play (Bottom Right).} 
\label{results_1}
\end{figure*}

\section{GPT-4V Empowered Embodied Task Planning}

Based on the aforementioned investigation into embodied tasks and LLMs, we developed an embodied task planning framework based on GPT-4V in this study and conducted evaluation experiments, as shown in Fig.~\ref{main}. The following section provides detailed information on the datasets, prompt design, and experimental results.

\subsection{Datasets}
To comprehensively evaluate the multimodal embodied task planning capabilities of GPT-4V, over 40 cases from 9 datasets \url{https://docs.google.com/spreadsheets/d/1rPBD77tk60AEIGZrGSODwyyzs5FgCU9Uz3h-3_t2A9g/edit#gid=0}are selected, focusing on manipulation and grasping. These actions are fundamental in instruction-following robotics, involving a variety of human instructions across diverse scenarios, such as kitchen pickups and tabletop rearrangements. The selected datasets are accessed through the Google Open X-Embodiment Dataset \cite{padalkar2023open}. 
In each case, video demonstrations and natural language instructions serve as inputs to assess GPT-4V as a robotic brain. This setup enables robust planning based on natural language instructions for generating robot actions.

\subsection{Prompt Design}
The design of prompts plays a crucial role in querying LLMs. A meticulously crafted prompt, rich in information and structured clearly, yields more precise and consistent outputs aligned with the given instructions. Here we update the text prompt from \cite{wake2023chatgpt} by incorporating images, creating a multimodal prompt that guides GPT-4V to produce robot task plans. The multimodal prompt consists of five parts:
\begin{itemize}
\item \textbf{System Role Explanation}: Specifies the task and the persona GPT-4V adopts in its responses.
\item \textbf{Predefined Action Pool}: A set of predefined robot actions from which GPT-4V can select and sequence to complete tasks step-by-step. To address vocabulary limitations, GPT-4V is prompted to create new actions if necessary.
\item \textbf{Example Output}: An example in JSON format to illustrate the expected output and ensure consistency.
\item \textbf{Case-by-Case Environment Image and Natural Language Instruction}: Includes the first frame extracted from the video demonstration as the environment image.
\item \textbf{Evaluation}: GPT-4V is tasked to assess the generated task plan against the ground truth video demonstration, scoring the plan based on its alignment with the video and providing an explanation.
\end{itemize}
The first three components are input as system messages for each query, while the last two as user messages vary according to the test data. The complete prompt is depicted in Fig.~\ref{prompt} of Appendix.




\begin{figure*}[t]
\centering
\includegraphics[width=1\textwidth]{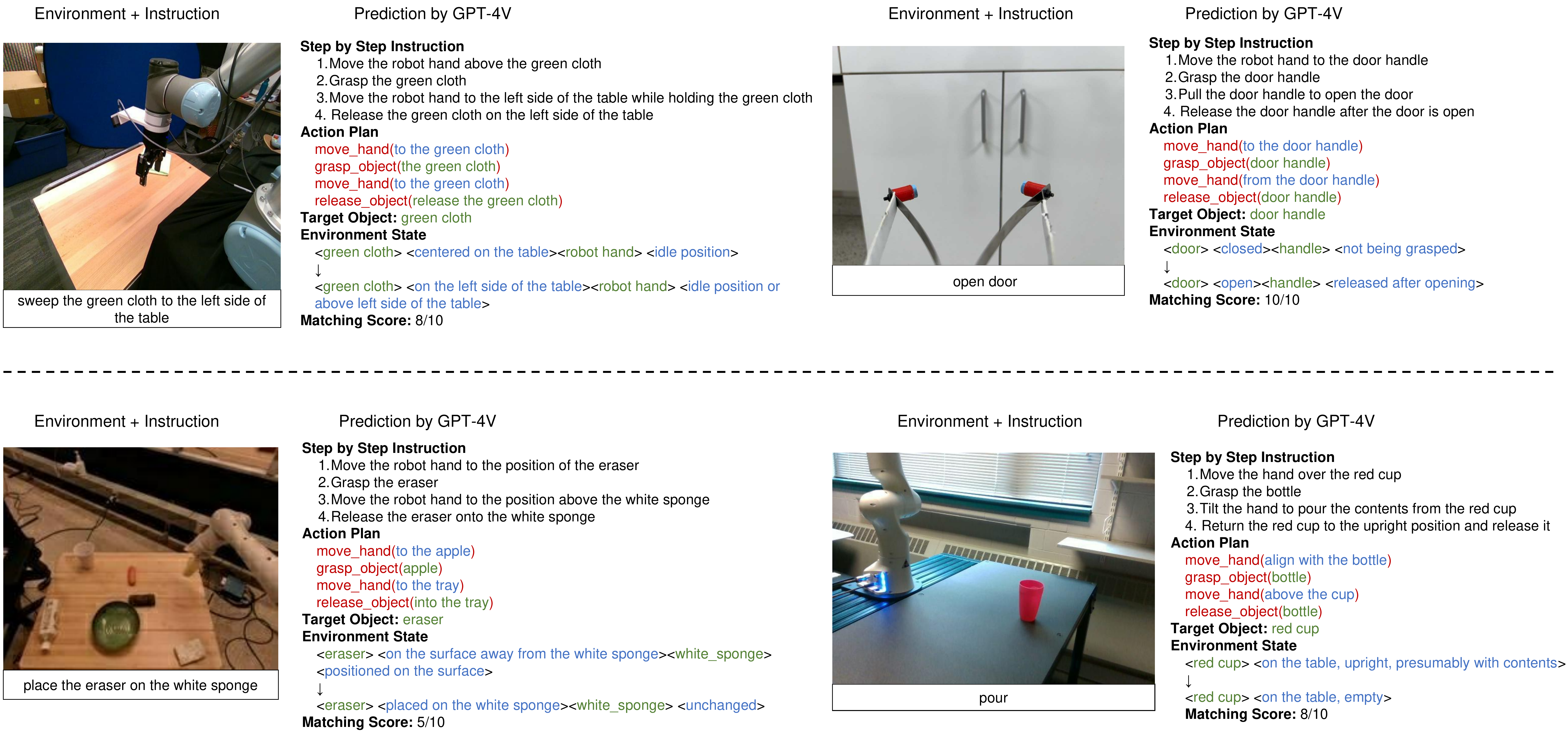}
\caption{Generated task plans for different datasets: Berkeley Autolab UR5 (Top Left), NYU VINN (Top Right), BC-Z (Bottom Left), and TOTO Benchmark (Bottom Right).} 
\label{results_2}
\end{figure*}

\section{Experimental Results}

In our experimental framework, the Large Language Models (LLMs) first generate step-by-step instructions tailored to the objectives of each robotic task. Subsequently, guided by these generated instructions, the model selects the most appropriate action from a predefined action pool and action objects to form the action plan for each step. After obtaining the instructions generated by the LLMs, we quantitatively evaluated the generated results by comparing them with the Ground-Truth instructions from the respective video dataset. Rigorous testing was conducted on 9 publicly available robot datasets, resulting in profound and insightful findings.

For instance, in the RT-1 Robot Action~\cite{brohan2022rt} dataset, as depicted in Fig.~\ref{results_1} top panel, the multi-modality LLMs accurately identified the target object and proficiently decomposed and executed the task. As shown in Fig.~\ref{results_1} top left, based on the given environment and instruction, the instructions generated by LLMs were as follows: 1) Move the hand to the target's location in the bottom drawer; 2) Grasp the target; 3) Move the hand with the target to the counter; 4) Release the target onto the counter. After providing detailed step-by-step textual instructions, the LLMs select from the action pool and lists a set of instructions and objects that comply with the current strategy. For example, "$move\_hand(bottom\ drawer)$" is the functional expression of the first textual instruction, facilitating subsequent direct usage of this action plan with the interface code controlling the robotic arm. Additionally, through the "Environment State" generated by the LLMs, it is evident that the models can effectively comprehend the changing spatial relationships of key objects in the environment after a series of operations. The "Matching Score" in Fig.~\ref{results_1} also demonstrates the model's precision.

In the aforementioned testing cases, the scenarios involved fewer objects and relatively concise and clear task instructions. Therefore, we further conducted tests involving semantically vague task descriptions and complex scenes. Fig.~\ref{results_1} middle left represents a test case from the QT-Opt dataset~\cite{kalashnikov2018qt}, where the instruction is simply "pick anything" without specifying any entities in the scene. From the results generated by LLMs, it produced a series of generalized instructions suitable for grasping any object, maintaining a high level of consistency with the ground truth. For complex scenes, as illustrated in Fig.~\ref{results_1} middle right, we tested an exemplary case from the Berkeley Bridge dataset~\cite{walke2023bridgedata}. The input instruction "Move the silver pot from in front of the red can, to next to the blue towel at the front edge of the table" involves multiple objects and their spatial relationships within the scene. Here, the LLMs not only grasped the task's purpose but also executed the task details adeptly, exemplifying their advanced image comprehension and logical reasoning abilities.


Further evidence of the LLMs' effectiveness in diverse and complex scenarios (including datasets~\cite{rosete2022tacorl,dass2023jacoplay,BerkeleyUR5Website,pari2021surprising,jang2021bc,zhou2023train}) is presented in Fig.~\ref{results_1} and Fig.~\ref{results_2}. Across these experiments, the LLMs demonstrated remarkable performance, even in tasks with intricate settings or specific requirements. Table~\ref{match_score} presents the average matching score, self-evaluated by GPT-4V across nine diverse datasets, indicating a consistently high level of agreement between the generated task plans and the ground truth demonstrations. This consolidates the validity of our approach and underscores the potent image understanding and logical reasoning capacities of multimodal LLMs in robotic task execution. Additional test results can be found in the Appendix.


\section{Limitation, Discussion and Future work}

We present an overview of integrating Large Language Models (LLMs) into robotic systems for various tasks and environments and evaluate GPT-4V in multimodal task planning. Although GPT-4V exhibits impressive multimodal reasoning and understanding capabilities as a robot brain for task planning, it faces several limitations: 1) The generated plans are homogenous, lacking in detailed embodiment and specific, robust designs to manage complex environments and tasks. 2) Current multimodal LLMs, such as GPT-4V and Google Gemini~\cite{team2023gemini}, necessitate carefully crafted, lengthy prompts to produce reliable outputs, which require domain expertise and extensive tricks. 3) The robot is constrained by predefined actions, limiting its executional freedom and robustness. 4) The closed-source nature of the GPT-4V API and associated time delays may impede embedded system development and real-time commercial applications. Future research should aim to address these challenges to develop more robust AGI robotic systems.

On the other hand, the advanced reasoning and vision-language understanding abilities exhibited by multimodal GPT-4V in robotics highlight the potential of LLM-centric AGI robotic systems. Moving forward, multimodal-LLM-centric AGI robots hold potential for application across various domains. In the realm of precision agriculture, these robots could supplant human labor in various labor-intensive tasks, especially in harvesting. This encompasses tasks like fruit picking and crop phenotyping~\cite{liu2023intelligent,liu2023dt}, which require advanced reasoning and precise action in the intricate environment of farms~\cite{lu2023agi}. In the healthcare domain, the critical need for safety and precision imposes greater demands on the perceptual and reasoning abilities of multimodal LLMs. This aspect is especially vital in robot-assisted screening and surgeries, where custom tasks tailored to individual needs are paramount~\cite{batailler2021mako}. Furthermore, leveraging contrastive learning models like CLIP~\cite{radford2021learning} to align brain signals with natural language suggests a pathway for developing Brain-Computer Interfaces (BCIs) in LLM-centric AGI robotic systems~\cite{duan2023dewave}. These systems could be capable of reading and interpreting human brain signals, such as EEG and fMRI, for self-planning and control in complex task completion~\cite{zhang2023noir,qiao2023brain}. This advancement could significantly bridge the gap in human-environment interaction and alleviate physical and cognitive labor.

\section{Conclusion}

In this paper, we have provided an overview of the integration of Large Language Models (LLMs) into various robotic systems and tasks. Our analysis reveals that LLMs demonstrate impressive reasoning, language understanding, and multimodal processing abilities that can significantly enhance robots' comprehension of instructions, environments, and required actions.

We evaluated the recently released GPT-4V model on over 30 cases across 9 datasets for embodied task planning. The results indicate that GPT-4V can effectively leverage natural language instructions and visual perceptions to generate detailed action plans to accomplish manipulation tasks. This suggests the viability of using multimodal LLMs as robotic brains for embodied intelligence.

However, some challenges remain to be addressed regarding model transparency, robustness, safety, and real-world applicability as we progress towards more practical and capable LLM-based AI systems. Specifically, the black-box nature of large neural models makes it difficult to fully understand their internal reasoning processes and failure modes. Additionally, bridging the gap between simulation and the real-world poses persisting difficulties in transferring policies without performance degradation. Extensive research is still needed to address these issues through techniques like standardized testing, adversarial training, policy adaptation methods, and safer model architectures. Accountability and oversight protocols for autonomous intelligent systems relying on LLMs also warrant thoughtful consideration. Overcoming these multifaceted challenges in a careful, ethical and socially responsible manner remains imperative as we advance progress in this domain.

As language models continue to accumulate extensive grounded knowledge from multimodal data, we anticipate rapid innovations in integrating them with robotics and simulation-based learning. This could enable intuitive development and validation of intelligent robots entirely in simulation using sim-to-real techniques before deployment. Such developments could profoundly enhance and transform how we build, test and deploy intelligent robotic systems.

Overall, the synergistic integration of natural language processing and robotics is a promising frontier filled with opportunities and challenges that warrant extensive future interdisciplinary research.

\bibliographystyle{IEEEtran}
\bibliography{mybib} 

\begin{figure*}[b]
\centering
\includegraphics[width=0.83\textwidth]{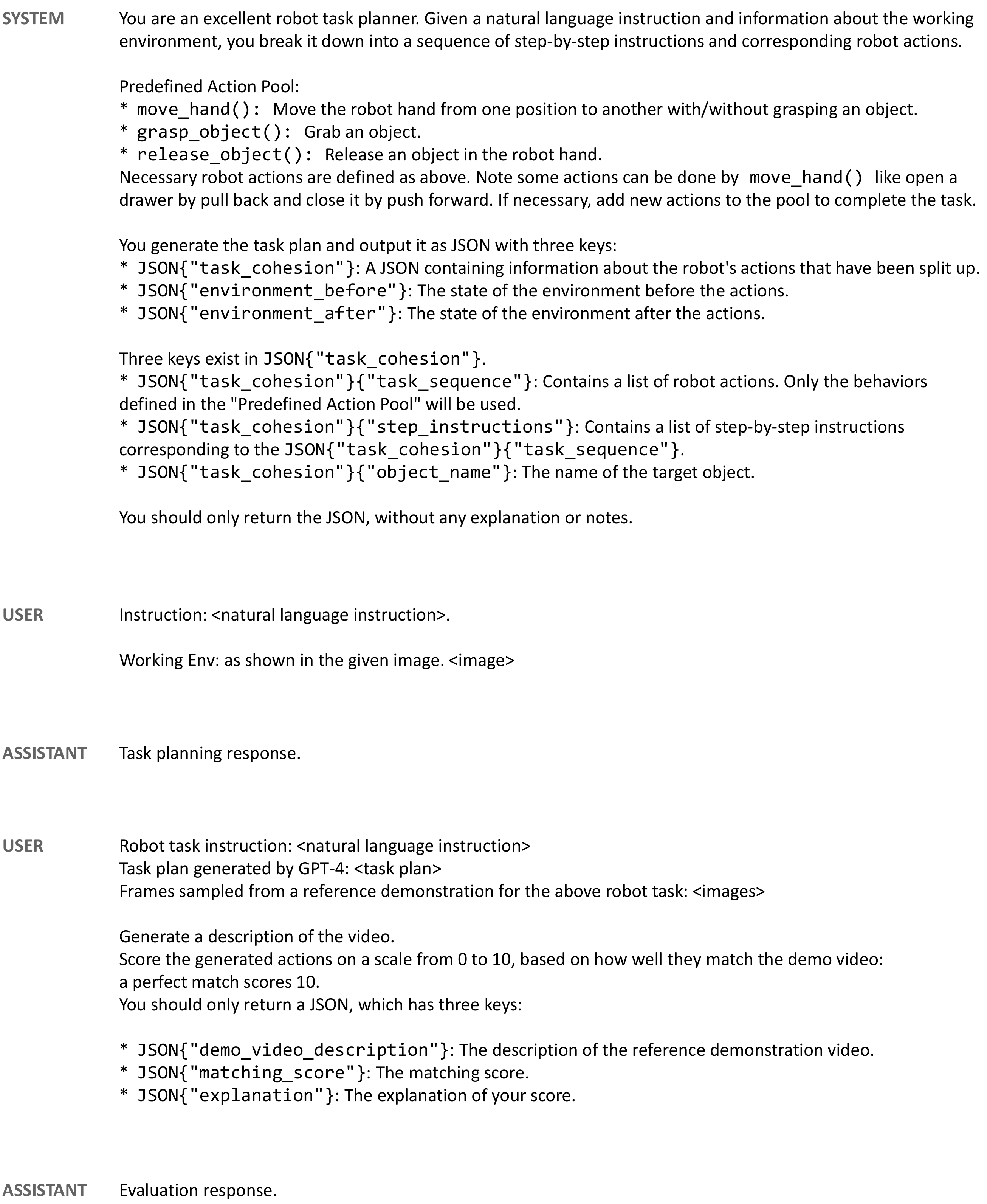}
\caption{Full prompts.} 
\label{prompt}
\end{figure*}
\begin{figure*}[h!]
\centering
\includegraphics[width=1\textwidth]{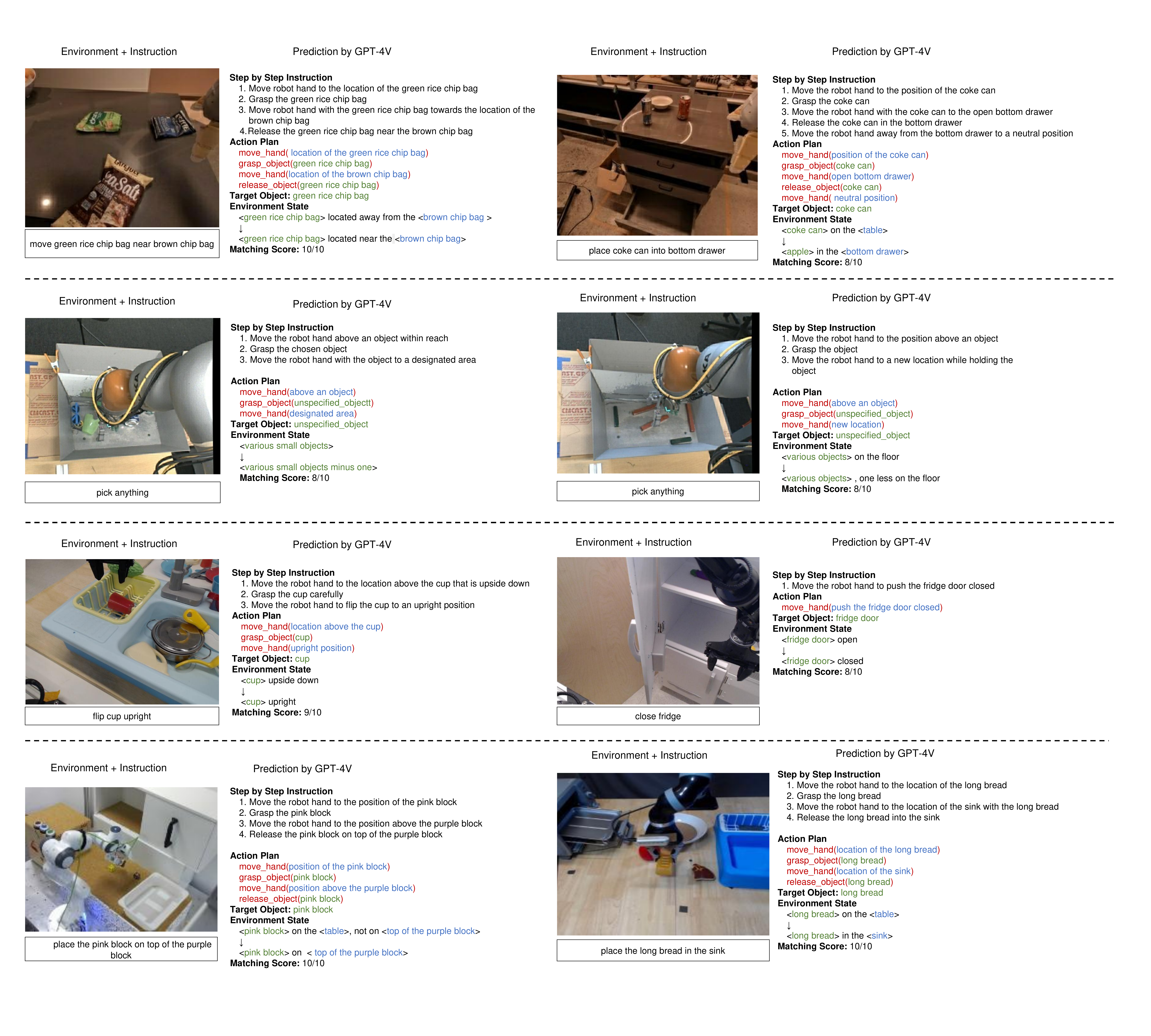}
\caption{Generated task plans for different datasets: RT-1 Robot Action (Top Panel), QT-Opt (Second Panel), Berkeley Bridge (Third Panel), Freiburg Franka Play (Bottom Left), and USC Jaco Play (Bottom Right).} 
\label{ap1}
\end{figure*}
\begin{figure*}[h!]
\centering
\includegraphics[width=1\textwidth]{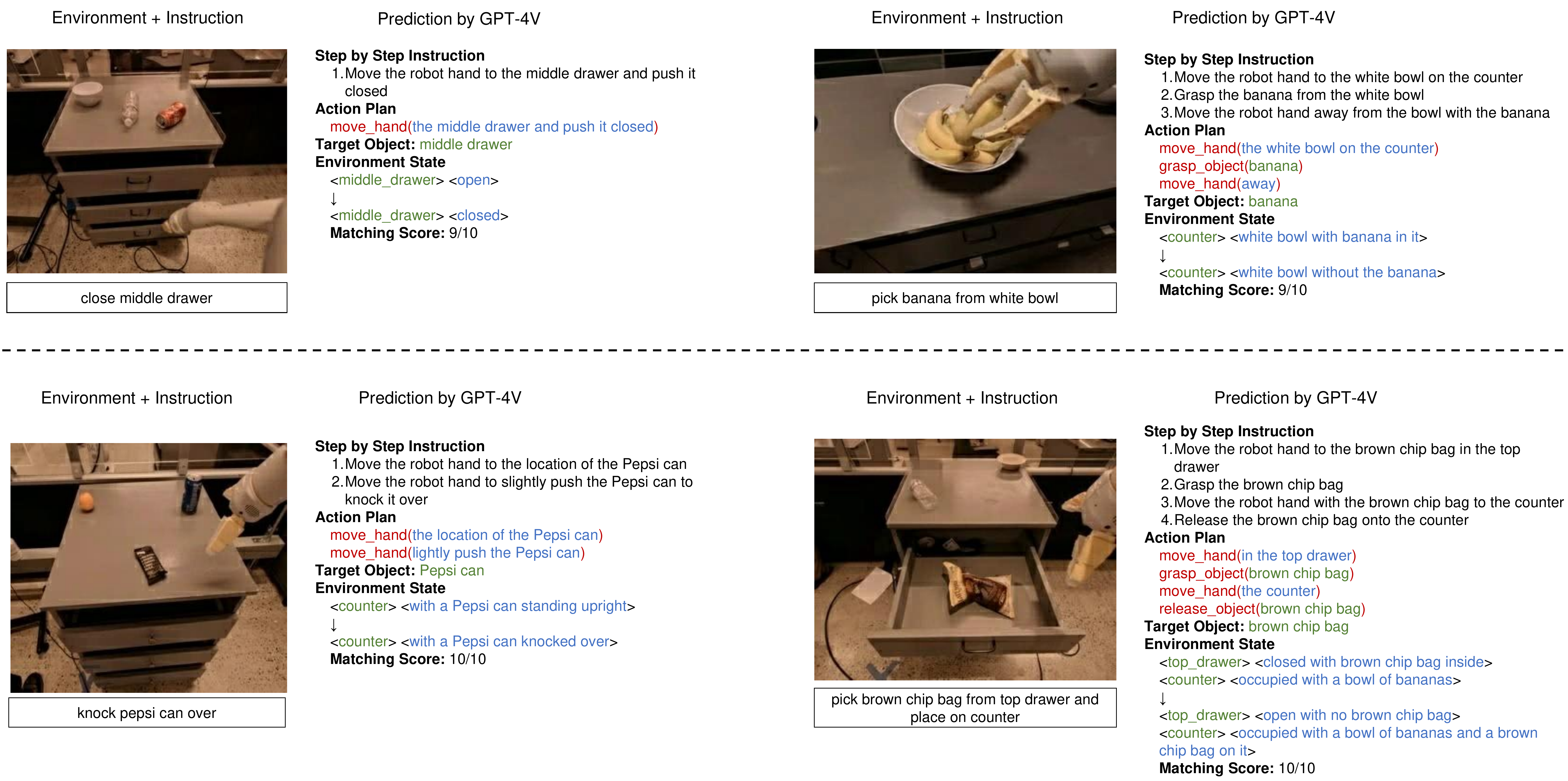}
\caption{Generated task plans for different datasets: RT-1 Robot Action.} 
\label{ap3}
\end{figure*}

\clearpage
\appendix
In this Appendix, the complete prompt (Fig.~\ref{prompt}) in our framework and additional experimental results are presented  (Fig.~\ref{ap1}$\sim$Fig.~\ref{ap3}).


\clearpage

\begin{figure*}[h!]
\centering
\includegraphics[width=1\textwidth]{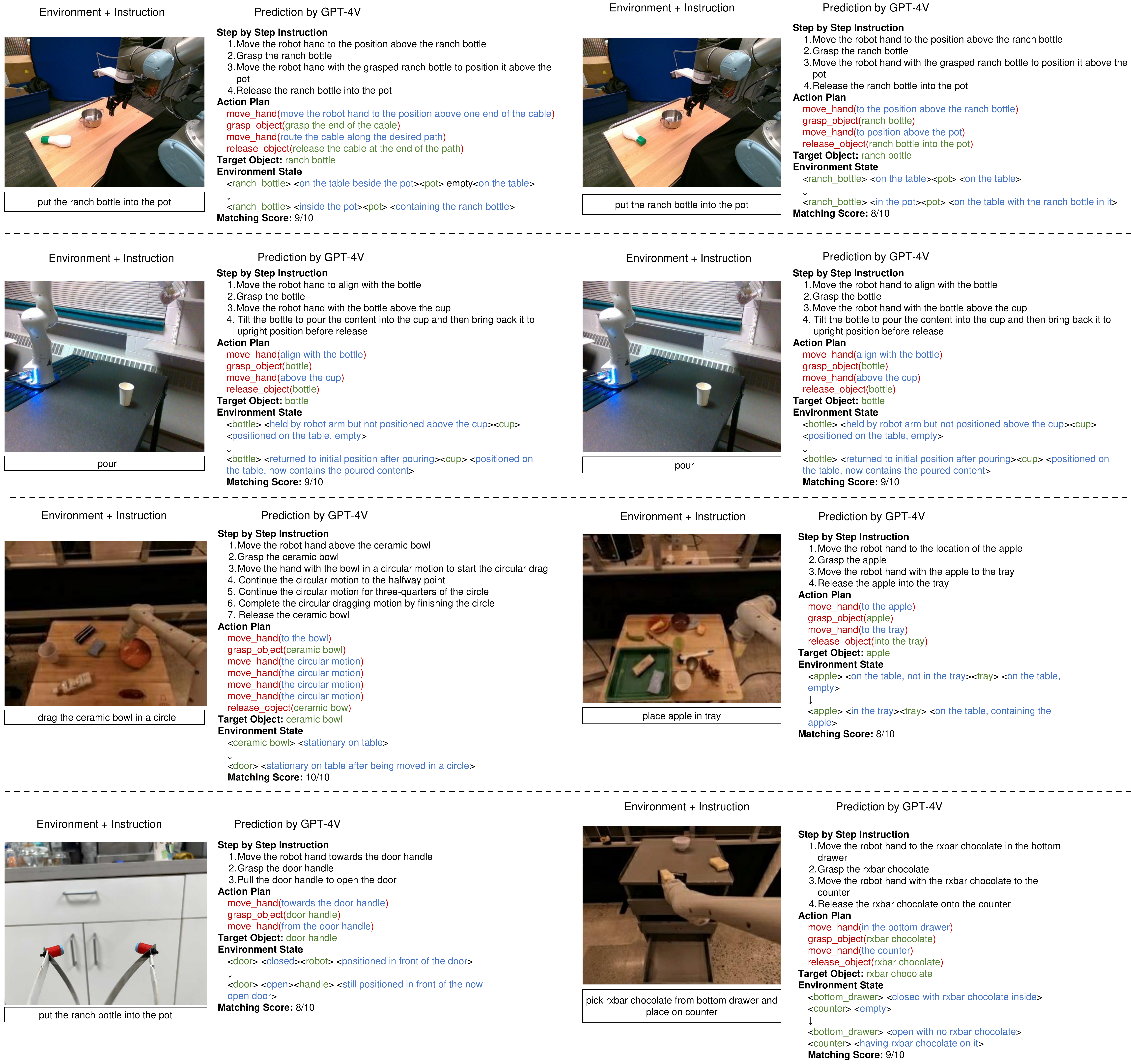}
\caption{Generated task plans for different datasets: Berkeley Autolab UR5 (Top Panel), TOTO Benchmar (Second Panel),  BC-Z (Third Panel), NYU VINN (Bottom Left). and RT-1 Robot Action (Bottom Right)} 
\label{ap2}
\end{figure*}

\vfill

\end{document}